\documentclass[10pt,twocolumn,letterpaper]{article}

\usepackage{cvpr}
\usepackage{times}
\usepackage{epsfig}
\usepackage{graphicx}
\usepackage{amsmath}
\usepackage{amssymb}
\usepackage{microtype}
\usepackage{color}
\usepackage{array}
\usepackage{hhline}
\usepackage{multirow}
\usepackage{caption}
\usepackage{subcaption}
\newcommand{\argmin}{\operatornamewithlimits{argmin}}
\usepackage{authblk}

\makeatletter
\renewcommand\AB@affilsepx{ \ \ \ \ \ \ \protect\Affilfont}
\makeatother

\usepackage[pagebackref=true,breaklinks=true,letterpaper=true,colorlinks,bookmarks=false]{hyperref}

\usepackage{enumitem}
\setitemize[0]{leftmargin=10pt}

\newenvironment{tight_itemize}{
\begin{itemize}[leftmargin=10pt]
  \setlength{\topsep}{0pt}
  \setlength{\itemsep}{2pt}
  \setlength{\parskip}{0pt}
  \setlength{\parsep}{0pt}
}{\end{itemize}}

\cvprfinalcopy 

\def\cvprPaperID{2317} 

\ifcvprfinal\fi

\begin{document}

\title{Deep Supervision with Shape Concepts for Occlusion-Aware 3D Object Parsing}
\author[1]{Chi Li}
\author[2]{M. Zeeshan Zia}
\author[2]{Quoc-Huy Tran}
\author[2]{Xiang Yu}
\author[1]{Gregory D. Hager}
\author[2,3]{Manmohan Chandraker}
\affil[1]{Johns Hopkins University}
\affil[2]{NEC Labs America}
\affil[3]{UC San Diego}
\maketitle

\begin{abstract}
Monocular 3D object parsing is highly desirable in various scenarios including occlusion reasoning and holistic scene interpretation. 
We present a deep convolutional neural network (CNN) architecture to localize semantic parts in 2D image and 3D space while inferring their visibility states, given a single RGB image. 
Our key insight is to exploit domain knowledge to regularize the network by deeply supervising its hidden layers, in order to sequentially infer intermediate concepts associated with the final task. 
To acquire training data in desired quantities with ground truth 3D shape and relevant concepts, we render 3D object CAD models to generate large-scale synthetic data and simulate challenging occlusion configurations between objects. 
We train the network only on synthetic data and demonstrate state-of-the-art performances on real image benchmarks including an extended version of KITTI, PASCAL VOC, PASCAL3D+ and IKEA for 2D and 3D keypoint localization and instance segmentation. 
The empirical results substantiate the utility of our deep supervision scheme by demonstrating effective transfer of knowledge from synthetic data to real images, resulting in less overfitting compared to standard end-to-end training.

\end{abstract}


\vspace{-0.1cm}
\section{Introduction}
\label{sec:introduction}
\vspace{-0.1cm}

The world around us is rich in structural regularity, particularly when we consider man-made objects such as cars or furniture. 
Studies in perception show that the human visual system imposes structure to reason about stimuli~\cite{Smith_1986}. 
Consequently, early work in computer vision studied perceptual organization as a fundamental precept for recognition and reconstruction \cite{Lowe_1985,Marr_1982}.
In particular, intermediate concepts like viewpoint are explored to aid complex perception tasks such as shape interpretation and mental rotation.
However, algorithms designed on these principles \cite{Mohan_Nevatia_1989,Sarkar_Soundararajan_2000} suffered from limitations in the face of real-world complexities because they relied on hand-crafted features (such as corners or edges) and hard-coded rules (such as junctions or parallelism).
In contrast, with the advent of convolutional neural networks (CNNs) in recent years, there has been tremendous progress in end-to-end trainable feature learning for object recognition, segmentation and reconstruction.

\begin{figure}[t]
  \centering
    \includegraphics[width=0.97\linewidth]{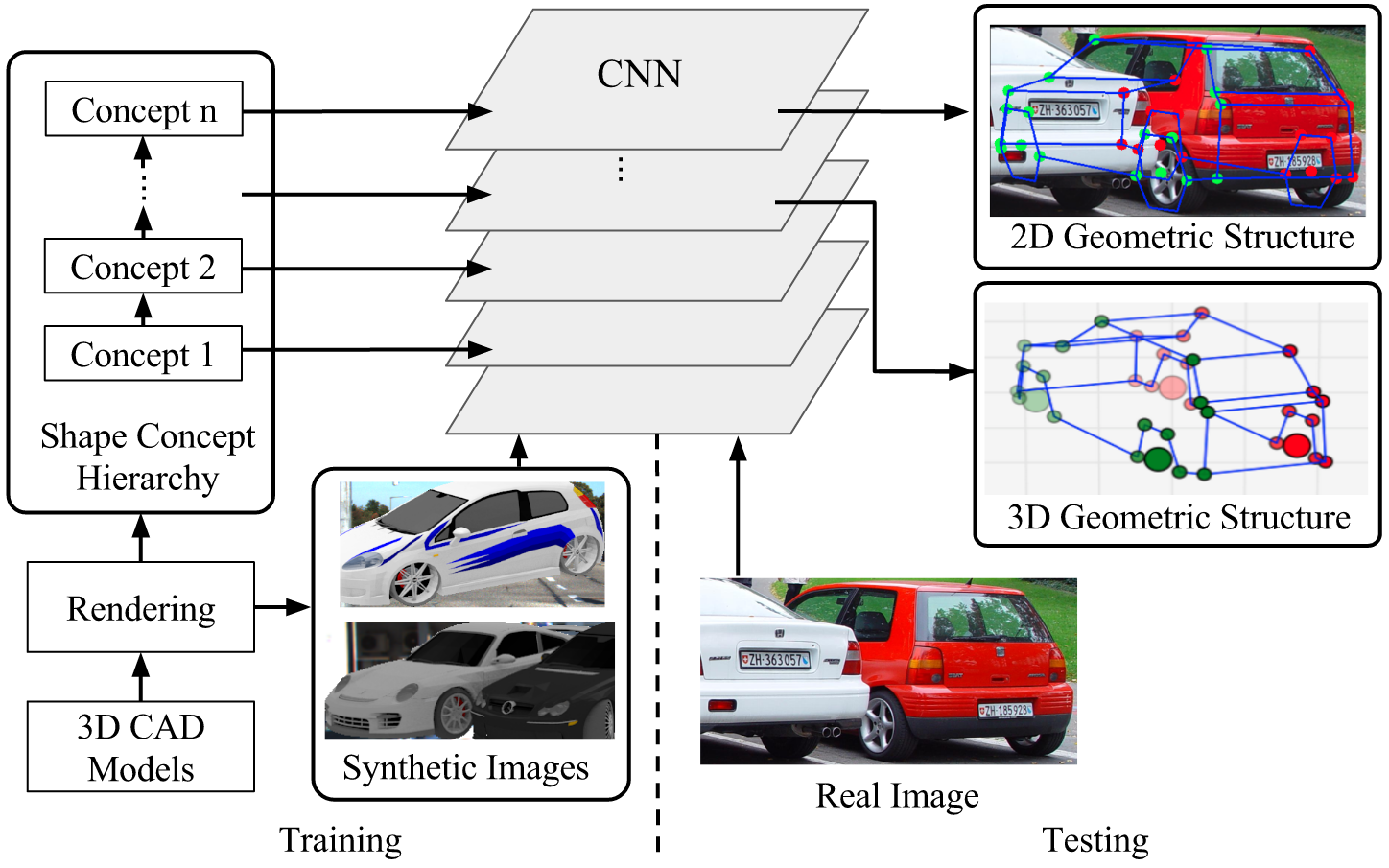}
  \caption{\small Overview of our approach. We use synthetic training images with intermediate shape concepts to deeply supervise the hidden layers of a CNN. At test time, given a single real image of an object, we demonstrate accurate localization of semantic parts in 2D and 3D, while being robust to intra-class appearance variations as well as occlusions.}
 \label{fig:intro}
 \vspace{-0.2cm}
\end{figure}

In this paper, we posit that it is advantageous to consider a middle ground, where we combine such early intuitions \cite{Marr_1982,Lowe_1985} on shape concepts with the discriminative power of modern CNNs to parse 2D/3D object geometry across intra-class appearance variations, including complex phenomena such as occlusions. 
Specifically, we demonstrate that intermediate shape concepts pertinent to 2D/3D shape understanding, such as pose and part visibility, can be applied to supervise intermediate layers of a CNN.
This allows greater accuracy in localizing the semantic elements of an object observed in a single image.

To illustrate this idea, we use 3D skeleton~\cite{torresani2003learning} as the shape representation, where semantically meaningful object parts (such as the wheels of a car) are represented by 3D keypoints and their connections define 3D structure of an object category. 
This representation is more efficient than 3D volumes \cite{choy20163d} or meshes \cite{zia09icar,tatarchenko2016multi,kar2015category,pol2016overcoming,kulkarni2014inverse,rezende2016unsupervised} in conveying the semantic information necessary for shape reasoning in applications such as autonomous driving. 

\begin{figure*}[t]
  \centering
    \includegraphics[width=1.0\linewidth]{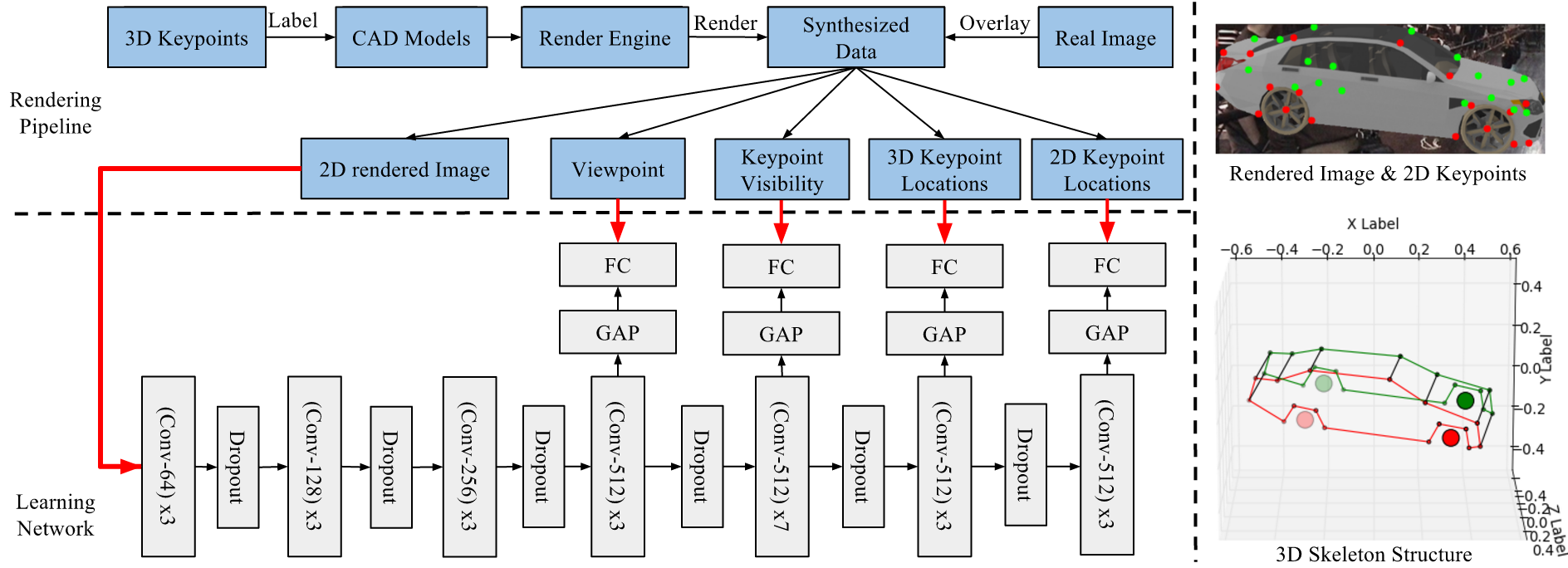}
  \caption{Visualization of our rendering pipeline (top-left), DISCO network (bottom-left), an example of rendered image and its annotations of 2D keypoints (top-right) as well as 3D skeleton (bottom-right). }
 \label{fig:pipe}
\end{figure*}

We introduce a novel CNN architecture which jointly models multiple shape concepts including object pose, keypoint locations and visibility in Section \ref{sec:method}. 
We first formulate the deep supervision framework by generalizing Deeply Supervised Nets~\cite{lee2015deeply} in Section \ref{sec:prob}.
In turn, Section \ref{sec:cnn} presents one particular network instance where we deeply supervise convolutional layers at different depths with intermediate shape concepts.
Further, instead of using expensive manual annotations, Section \ref{sec:data} proposes to render 3D CAD models to create synthetic images with concept labels and simulate the challenging occlusion configurations for robust occlusion reasoning.
Figure \ref{fig:intro} introduces our framework and Figure \ref{fig:pipe} illustrates a particular instance of deeply supervised CNN using shape concepts.
We denote our network as ``DISCO'' short for Deep supervision with Intermediate Shape COncepts.

At test time, DISCO trained on only synthetic images generalizes well to real images. 
In particular, it empirically outperforms single-task architectures without supervision for intermediate shape concepts and multitask networks which impose supervision of all the concepts at the top layer.
This observation demonstrates the intimacy of shape concepts for 3D object parsing, despite the fact that we ignore aspects of photorealism such as material and illumination in our rendered training data.
In Section \ref{sec:exp}, we quantitatively demonstrate significant improvements over prior state-of-the-art for 2D keypoint and 3D structure prediction on PASCAL VOC, PASCAL3D+\cite{xiang2014beyond}, IKEA\cite{lim2013parsing} and an extended KITTI \cite{geiger2012we} dataset (KITTI-3D).

We note that most existing approaches \cite{zia09icar,zia2013detailed,kar2015category,kulkarni2014inverse,wu2015learning,zhou2016learning} estimate 3D geometry by comparing projections of parameterized shape models with separately predicted 2D patterns, such as keypoint locations or heat maps. This makes prior methods sensitive to partial view ambiguity \cite{lee1985determination} and incorrect 2D structure predictions. 
Moreover, scarce 3D annotations for real images further limit their performance. 
In contrast, we make the following novel contributions to alleviate those problems:
\vspace{-0.2cm}
\begin{tight_itemize}
\item We demonstrate the utility of rendered data with access to intermediate shape concepts. In addition, we model occlusions by appropriately rendering multiple object configurations, which presents a novel way of exploiting 3D CAD data for realistic scene interpretation.
\item We apply intermediate shape concepts to deeply supervise the hidden layers of a CNN. This approach exhibits the better generalization from synthetic to real images than the standard end-to-end training.
\item Our method achieves state-of-the-art performance on 2D/3D semantic part localization under occlusion and large appearance changes on several public benchmarks.
\end{tight_itemize}

\section{Related Work}
\label{sec:related}

\vspace{-0.15cm}
\paragraph{3D Skeleton Estimation} 
This class of work models 3D shape as a linear combination of shape bases and optimizes basis coefficients to fit computed 2D patterns such as heat maps \cite{zhou2016learning} or object part locations \cite{zia2013detailed}. The single image 3D interpreter network (3D-INN) \cite{wu2016single} presents a sophisticated CNN architecture to estimate a 3D skeleton based only on detected visible 2D joints. The training of 3D-INN is not jointly optimized for 2D and 3D keypoint localization. Further, the decoupling of 3D structure from rich object appearance leads to partial view ambiguity and thus 3D prediction errors.


\vspace{-0.5cm}
\paragraph{3D Reconstruction}
A generative inverse graphics model is formulated by~\cite{kulkarni2014inverse} for 3D mesh reconstruction by matching mesh proposals to extracted 2D contours. 
Recently, given a single image, autoencoders have been exploited for 2D image rendering \cite{dosovitskiy2016learning}, multi-view mesh reconstruction \cite{tatarchenko2016multi} and 3D shape regression under occlusion \cite{pol2016overcoming}. The encoder network learns to invert the rendering process to recognize 3D attributes such as object pose. However, methods such as \cite{tatarchenko2016multi,pol2016overcoming} are quantitatively evaluated only on synthetic data and seem to achieve limited generalization to real images. Other works such as \cite{kar2015category} formulate an energy-based optimization framework involving appearance, keypoint and normal consistency for dense 3D mesh reconstruction, but require both 2D keypoint and object segmentation annotations on real images for training. Volumetric frameworks with either discriminative \cite{choy20163d} or generative \cite{rezende2016unsupervised} modeling infer a 3D shape distribution in voxel grids given one or multiple images of the same object. However, due to the highly redundant nature of voxel grid representations, they are limited to low resolutions up to $32$x$32$x$32$ for now. Lastly, 3D voxel examplars \cite{xiang2015data} jointly recognize the 3D shape and occlusion pattern by template matching~\cite{pepik13cvpr}, which is not scalable to more object types and complex shapes.    


\vspace{-0.5cm}
\paragraph{3D Model Retrieval and Alignment}
This line of work estimates 3D object structure by retrieving the closest object CAD model and performing alignment, using 2D images \cite{zia09icar,aubry2014seeing,lim2014fpm,massa2015deep,xiang2014beyond} and RGB-D data \cite{bansal2016marr,gupta2015inferring}. Unfortunately, limited number of CAD models can not represent all instances in one object category, despite explicit shape modeling~\cite{zia09icar}.
Further, the retrieval step is slow for a large CAD dataset and the alignment is sensitive to error in estimated pose.

\vspace{-0.5cm}
\paragraph{Pose Estimation and 2D Keypoint Detection}
``Render for CNN''~\cite{su2015render} synthesizes 3D CAD model views  as additional training data besides real images for object viewpoint estimation. We extend this rendering pipeline to support object keypoint prediction and model occlusion. Viewpoint prediction is utilized in \cite{tulsiani2015viewpoints} to significantly boost the  performance of 2D landmark localization. Recent work such as DDN~\cite{yu2016deep} optimizes deformation coefficients based on the PCA representation of 2D keypoints to achieve state-of-the-art performance on face and human body. Dense feature matching approaches which exploit top-down object category knowledge \cite{kanazawa2016warpnet,zhou2016learning} also obtain recent successes, but our method yields better results.


\section{Deep Supervision with Shape Concepts}
\label{sec:method}

In the following, we introduce a novel CNN architecture for 3D shape parsing which incorporates constraints through intermediate shape concepts such as object pose, keypoint locations, and visibility information.
Our goal is to infer, from a single view (RGB image) of the object, the locations of keypoints in 2D and 3D spaces and their visibility.
We motivate our deep supervision scheme in Section \ref{sec:prob}. 
Subsequently, we present the network architecture in Section \ref{sec:cnn} which exploits synthetic data generated from the rendering pipeline detailed in Section \ref{sec:data}.

\subsection{Deep Supervision}
\label{sec:prob}
Our approach draws inspiration from Deeply Supervised Nets (DSN) \cite{lee2015deeply}. 
However, whereas DSN supervises each layer by the final label to accelerate training convergence, we sequentially apply deep supervision on intermediate concepts intrinsic to the ultimate task, in order to regularize the network for better generalization.

Let $\mathcal{Z}=\{(x,y)\}$ represent the training set with pairs of input $x$ and labels $y$ for a supervised learning task. The associated optimization problem for a multi-layer CNN is:
\begin{equation}
W^*=\argmin_W \sum_{(x,y)\in\mathcal{Z}} l(y,f(x,W)) 
\label{eq:obj}
\end{equation}
\noindent
where $l(.,.)$ is a problem specific loss, $W=\{W_1,...,W_N\}$ stands for the weights of N layers, and function $f$ is defined based on the network structure.
In practice, the optimal solution $\widehat{W}^*$ may suffer from overfitting. That is, given a new population of data $\mathcal{Z'}$, the performance of $f(\cdot,W)$ on $\mathcal{Z'}$ is substantially lower than on $\mathcal{Z}$. This is particularly the case when, for example, we train on synthetic data but test on real data.

One way to address the overtraining is through regularization which biases the network to incrementally reproduce physical quantities that are relevant to the final answer. 
For example, object pose is an indispensable element to predict 3D keypoint locations. 
Intuitively, the idea is to prefer solutions that reflect the underlying physical structure of the problem which is entangled in the original training set.
Because deeper layers in CNNs represent more complex concepts due to growing size of receptive fields and more non-linear transformations stacked along the way, we may realize our intuition by explicitly enforcing that hidden layers yield a sequence of known intermediate concepts with growing complexity towards the final task.

To this end, we define the augmented training set $\mathcal{A}=\{(x,\{y_1,...,y_N\})\}$ with additional supervisory signals $\{y_1,...,y_{N-1}\}$. 
Further, we denote $W_{1:i} = \{W_1, \ldots, W_i\}$ as the weights for the first $i$ layers of the CNN and $h_i = f(\cdot, W_{1:i})$ as the activation map of layer $i$.
We now extend \eqref{eq:obj} to the additional training signals $y_i$ by introducing functions $y_i=g(h_i,v_i)$ parameterized by the weight $v_i$.
Letting $V = \{v_1, \ldots, v_{N-1}\}$, we can now write a new objective trained over
$\mathcal{A}$: 
\begin{equation}
\widehat{W}^*,\widehat{V}^* = \argmin_{W,V} \!\!\!\! \sum_{(x,\{y_i\})\in\widehat{\mathcal{A}}} \sum_{i=1}^N \lambda_i l_i(y_i,g(f(x,W_{1:i}),v_i))
\label{eq:approx}
\end{equation}
\noindent
The above objective can be optimized by simultaneously backpropagating the errors of all supervisory signals scaled by $\lambda_i$ on each $l_i$ to $W_{1:i}$.  From the perspective of the original problem, new constraints through $y_i$ act as additional regularization on the hidden layers, thus biasing the network toward solutions that, as we empirically show in Section~\ref{sec:exp}, exhibit better generalization than solutions to \eqref{eq:obj}.

\subsection{Network Architecture}
\label{sec:cnn}

To set up \eqref{eq:approx}, we must first choose a sequence of necessary conditions for 2D/3D keypoint prediction with growing complexity as intermediate shape concepts. 
We have chosen, in order, (1) object viewpoint, (2) keypoint visibility, (3) 3D keypoint locations and (4) full set of 2D keypoint locations regardless of the visibility, inspired by early intuitions on perceptual organization~\cite{Marr_1982,Lowe_1985}. 
We impose this sequence of intermediate concepts to deeply supervise the network at certain depths as shown in Fig.~\ref{fig:pipe} and minimize four intermediate losses $l_i$ in \eqref{eq:approx}, with other losses removed.

Our network resembles the VGG network \cite{simonyan2014very} and consists of deeply stacked $3\times 3$ convolutional layers.
Unlike VGG, we remove local spatial pooling and couple each convolutional layer with batch normalization \cite{ioffe2015batch} and ReLU, which defines the $f(x,W_{1:i})$ in \eqref{eq:approx}. 
This is motivated by the intuition that spatial pooling leads to the loss of spatial information.
Further, $g(h_i,v_i)$ is constructed with one global average pooling (GAP) layer followed by one fully connected (FC) layer with $512$ neurons, which is different from stacked FC layers in VGG.
In Sec. \ref{sec:kitti3d}, we empirically show that these two changes are critical to significantly improve the performance of VGG like networks for 2D/3D landmark localization.

To further reduce the issue of over-fitting, we deploy dropout \cite{krizhevsky2012imagenet} between the hidden convolutional layers. 
At layers 4,8,12, we perform the downsampling using convolution layers with stride 2.
Fig. \ref{fig:pipe} (bottom-left) illustrates our network architecture in detail.
We use L2 loss at all points of supervision.
``(Conv-A)xB'' means A stacked convolutional layers with filters of size BxB.
We deploy $25$ convolutional layers in total.

In experiments, we only consider the azimuth angle of the object viewpoint with respect to a canonical pose. 
We further discretize the azimuth angle into $M$ bins and regress it to a one-hot encoding (the entry corresponding to the predicted discretized pose is set to $1$ and all others to $0$).  
Keypoint visibility is also represented by a binary vector with $1$ indicating occluded state of a keypoint.
2D keypoint locations are normalized to $[0,1]$ with the image size along the width and height dimensions.
We center 3D keypoint coordinates of a CAD model at the origin and scale them to set the longest dimension (along X,Y,Z) to unit length. 
CAD models are assumed to be aligned along the principal coordinate axes, and registered to a canonical pose, as is the case for ShapeNet~\cite{chang2015shapenet} dataset.
During training, each loss is backpropagated to train the network jointly.

\subsection{Synthetic Data Generation}
\label{sec:data}

\begin{figure}[t]
  \centering
    \includegraphics[width=1.0\linewidth]{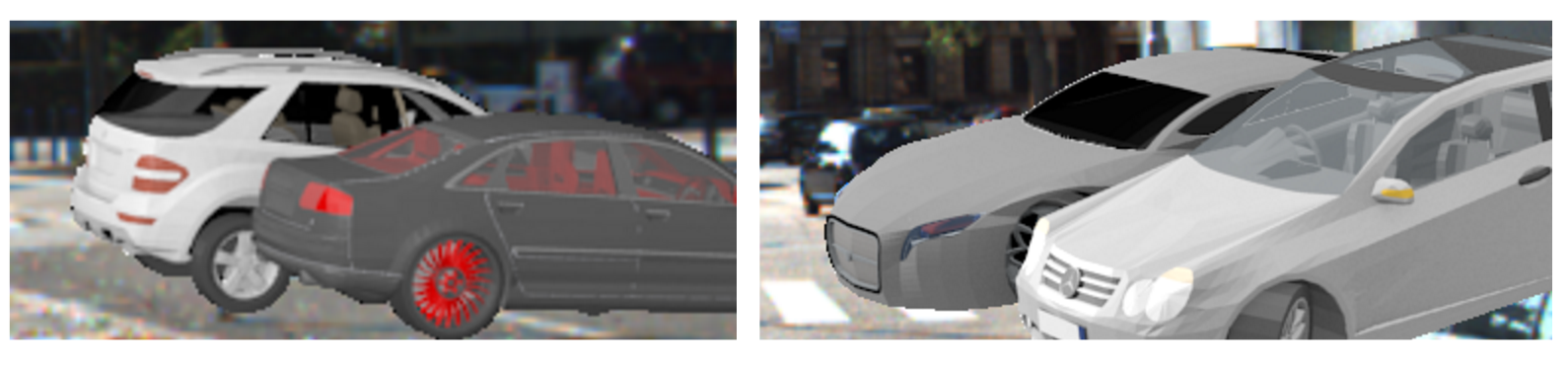}
  \caption{Examples of synthesized training images for simulating the object-object occlusion.}
 \label{fig:occ_examples}
\end{figure}

Unsurprisingly, our approach needs a large amount of training data because it is based on deep CNNs and involves more fine-grained labels than other visual tasks such as object classification. 
Furthermore, we aim for the method to work with occluded test cases.
Therefore, we need to generate training examples that are representative of realistic occlusion configurations caused by multiple objects in close proximity as well as image boundary truncations.
To obtain such large-scale training data, we extend the data generation pipeline of ``Render for CNN'' \cite{su2015render} with 2D/3D landmarks and visibility information. 

An overview of the rendering process is shown in the upper-left of Fig. \ref{fig:pipe}. 
We pick a small subset of CAD models from ShapeNet~\cite{chang2015shapenet} for a given object category and manually annotate 3D keypoints on each CAD model.
Next, we render each CAD model using the open-source tool Blender while randomly sampling rendering parameters from a uniform 
distribution including camera viewpoint, number/strength of light sources, and surface gloss reflection.
Finally, we overlay the rendered images on real image backgrounds to avoid over-fitting to synthetic data \cite{su2015render}.
We crop the object from each rendered image and extract the object viewpoint, 2D/3D keypoint locations and their visibility states from the render engine as the training labels. 
In Fig. \ref{fig:pipe}, we show an example of rendering and its 2D/3D annotations.

To model multi-object occlusion, we randomly select two different object instances and place them close to each other without overlapping in 3D space. 
During rendering, we compute the occlusion ratio of each instance by calculating the fraction of visible 2D area versus the complete 2D projection of CAD model. 
Keypoint visibility is computed by ray-tracing. 
We select instances with occlusion ratios ranging from $0.4$ to $0.9$. 
Fig.~\ref{fig:occ_examples} shows two representative training examples where cars are occluded by other nearby cars. 
For truncation, we randomly select two image boundaries (left, right, top, or bottom) of the object and shift them by $[0,0.3]$ of the image size along that dimension.

\section{Experiment}
\label{sec:exp}

\vspace{-0.1cm}
\paragraph{Dataset and metrics}
We empirically demonstrate competitive or superior performance compared to several state-of-the-art methods, on a number of public datasets: PASCAL VOC (Sec.~\ref{sec:voc}), PASCAL3D+ \cite{xiang2014beyond} (Sec.~\ref{sec:pascal3d}) and IKEA \cite{lim2013parsing} (Sec.~\ref{sec:ikea}). In addition, we evaluate our method on KITTI-3D where we generate 3D keypoint annotations on a subset of car images from KITTI dataset \cite{geiger2012we}. For training, we select $472$ cars, $80$ sofa and $80$ chair CAD models from ShapeNet \cite{chang2015shapenet}. 
Each car model is annotated with $36$ keypoints~\cite{zia2013detailed} and each sofa or chair model is labeled with $14$ keypoints~\cite{xiang2014beyond} \footnote{We use 10 chair keypoints consistent with \cite{wu2016single} for evaluation on IKEA.}.
We synthesize 600k car images including occluded instances and 200k images of fully visible furniture (chair+sofa).
We select rendered images of 5 CAD models from each object category as the validation set.

We use PCK and APK metrics~\cite{yang2011articulated} to evaluate the accuracy of 2D keypoint localization. 
A 2D keypoint prediction is correct when it lies within a specified radius $\alpha*L$ of the ground truth, where $L$ is the larger dimension of the image with $0 < \alpha < 1$. 
PCK is the percentage of correct keypoint predictions given the object location and keypoint visibility.
APK is the mean average precision of keypoint detection computed by associating each estimated keypoint with a confidence score. 
In our experiments, we use the regressed values of keypoint visibility as confidence scores. We extend 2D PCK and APK metrics to 3D by defining a correct 3D keypoint prediction whose euclidean distance to the ground truth is less than $\alpha$ in normalized coordinates.

\vspace{-0.6cm}
\paragraph{Training details}
We set loss weights of object pose to $0.1$ and others to $1$.
We use stochastic gradient descent with momentum $0.9$ to train the proposed CNN from scratch. 
The learning rate starts at $0.01$ and decreases by one-tenth when the validation error reaches a plateau. 
The weight decay is set to $0.0001$ and the input image size is $64$x$64$. 
The network is initialized following \cite{glorot2010understanding} and the batch size is $100$. 
For car model training, we form each batch using a mixture of fully visible, truncated and occluded cars, numbering $50$, $20$ and $30$, respectively. 
For furniture, each batch consists of $100$ images of chair and sofa mixed with random ratios. 
The network is implemented with Caffe~\cite{jia2014caffe}.

\subsection{KITTI-3D}
\label{sec:kitti3d}
\begin{table*}[t]
\small
\centering
\begin{tabular}{|  >{\centering\arraybackslash}m{0.98in} |  >{\centering\arraybackslash}m{0.6in} |  >{\centering\arraybackslash}m{0.6in} | >{\centering\arraybackslash}m{0.9in} | >{\centering\arraybackslash}m{0.6in} | >{\centering\arraybackslash}m{0.6in} || >{\centering\arraybackslash}m{0.6in} | >{\centering\arraybackslash}m{0.6in} | >{\centering\arraybackslash}m{0.6in} |}
\hline
\multirow{2}{*}{Method}	&	\multicolumn{5}{c||}{2D} & 3D & 3D-yaw	\\
\hhline{|~|-|-|-|-|-|-|-|} & Full & Truncation & Multi-Car Occ & Other Occ & All & Full &Full \\
\hline
DDN~\cite{yu2016deep} & 67.6 & 27.2 & 40.7 & 45.0 & 45.1 & \multicolumn{2}{c|}{NA} \\
\hline
WN-gt-yaw*~\cite{kanazawa2016warpnet}  & 88.0 & 76.0 & 81.0 & 82.7 & 82.0 & \multicolumn{2}{c|}{NA} \\
\hline
Zia et al.~\cite{zia2013detailed} & 73.6 & \multicolumn{4}{c||}{NA} & 73.5 & 7.3 \\
\hhline{|=|=|=|=|=|=||=|=|}
DSN-2D     	  & 45.2 & 48.4 & 31.7 & 24.8 & 37.5 & \multicolumn{2}{c|}{NA} \\
\hline
DSN-3D     	  & \multicolumn{5}{c||}{NA} & 68.3 & 12.5 \\
\hline
plain-2D      & 88.4 & 62.6 & 72.4 & 71.3 & 73.7 & \multicolumn{2}{c|}{NA} \\
\hline
plain-3D      & \multicolumn{5}{c||}{NA} & 90.6 & 6.5 \\
\hline
plain-all     & 90.8 & 72.6 & 78.9 & 80.2 & 80.6 & 92.9 & 3.9 \\
\hline
DISCO-3D-2D	  & 90.1 & 71.3 & 79.4 & 82.0 & 80.7 & 94.3 & 3.1 \\ 
\hline
DISCO-vis-3D-2D & 92.3 & 75.7 & 81.0 & 83.4 & 83.4 & 95.2 & 2.3 \\
\hline
DISCO-(3D-vis) & 87.8 & 76.1 & 71.0 & 68.3 & 75.8 & 89.7 & 3.6 \\
\hline
DISCO-reverse & 30.0 & 32.6 & 22.3 & 16.8 & 25.4 & 49.0 & 22.8 \\
\hline
DISCO-Vgg     & 83.5 & 59.4 & 70.1 & 63.1 & 69.0 & 89.7 & 6.8 \\ 
\hline
DISCO	 	  & \textbf{93.1} & \textbf{78.5} & \textbf{82.9} & \textbf{85.3} & \textbf{85.0} & \textbf{95.3} & \textbf{2.2} \\ 
\hline
\end{tabular}
\caption{\small PCK[$\alpha=0.1$] accuracies (\%) of different methods for 2D and 3D keypoint localization on KITTI-3D dataset. WN-gt-yaw~\cite{kanazawa2016warpnet} uses groundtruth pose of the test car. The bold numbers indicate the best results.}
\label{tab:car3d}
\end{table*}

We create a new KITTI-3D dataset for evaluation, using 2D keypoint annotations of $2040$ KITTI~\cite{geiger2012we} car instances provided by Zia et al.~\cite{zia2013detailed} and further labeling each car image with occlusion type and 3D keypoint locations.
We define four occlusion types: no occlusion (or fully visible cars), truncation, multi-car occlusion (the target car is occluded by other cars) and occlusion caused by other objects. The number of images for each type is $788$, $436$, $696$ and $120$, respectively. To obtain 3D groundtruth, we fit a PCA model trained on the 3D keypoint annotations on CAD data, by minimizing the 2D projection error for the known 2D landmarks.
We only provide 3D keypoint labels for fully visible cars because the occluded or truncated cars do not contain enough visible 2D keypoints for precise 3D alignment. We refer the readers to the supplementary material for more details about the 3D annotation and some labeled examples in KITTI-3D. 

Table \ref{tab:car3d} reports PCK accuracies for current state-of-the-art methods including DDN \cite{yu2016deep} and WarpNet \cite{kanazawa2016warpnet} for 2D keypoint localization and Zia et al.~\cite{zia2013detailed} for 3D structure prediction\footnote{We cannot report Zia et al.\cite{zia2013detailed} on occlusion categories because only a subset of images has valid results in those classes.}.
We use source codes for these methods provided by the respective authors.
Further, we enhance WarpNet (denoted as WN-gt-yaw) by using groundtruth poses of test images to retrieve $30$ labeled synthetic car images for landmark transfer, using median landmark locations as result.
We observe that DISCO outperforms these competitors on all occlusion types. 

We also perform a detailed ablative study on DISCO architecture.
First, we incrementally remove the deep supervision used in DISCO one by one. 
DISCO-vis-3D-2D, DISCO-3D-2D, plain-3D and plain-2D are networks without pose, pose+visibility, pose+visibility+2D and pose+visibility+3D, respectively.
We observe a monotonically decreasing trend of 2D and 3D accuracies: plain-2D or plain-3D $<$ DISCO-3D-2D $<$ DISCO-vis-3D-2D $<$ DISCO. 
Next, if we switch 3D and visibility supervision (DISCO-(3D-vis)), reverse the entire supervision order (DISCO-reverse) or move all supervision to the last convolutional layer (plain-all), the performance of these variants drop compared to DISCO.
In particular, DISCO-reverse decreases PCK by nearly $60\%$. 
We also find DISCO is much better than DSN-2D and DSN-3D which replace all intermediate supervisions with 2D and 3D labels, respectively.
This indicates that the deep supervision achieves better regularization during training by coupling the sequential structure of shape concepts with the feedforward nature of a CNN .
With the proposed order held fixed, when we deploy more than 10 layers before the first supervision and more than 2 layers between every two consecutive concepts, the performance of DISCO only varies by at most 2\% relative to the reported ones.
Finally, DISCO-VGG performs worse than DISCO by $16.0\%$ on 2D-All and $5.6\%$ on 3D-Full, which confirms our intuition to remove local spatial pooling and adopt global average pooling.

We also evaluate DISCO on detection bounding boxes computed from RCNN~\cite{girshick2014rich} with IoU$>0.7$ to the groundtruth of KITTI-3D.
The PCK accuracies by DISCO on 2D-All and 3D-Full are $88.3\%$ and $95.5\%$ respectively, which are even better than for true bounding boxes in Table \ref{tab:car3d}. 
It can be attributed to the fact that 2D groundtruth locations in KITTI do not tightly bound the object areas because they are only the projections of 3D groundtruth bounding boxes.
This result shows that DISCO is robust to imprecise 2D bounding boxes.
We refer readers to more numerical details in the supplementary material.
Last, we train DISCO over fully visible cars only and find that the accuracies of 2D keypoint localization decrease by $1.3\%$ on fully visible data, $24.9\%$ on truncated cases and $15.9\%$ on multi-car+other occluded cars. 
This indicates that the occlusion patterns learned on simulated occluded data is generalizable to real images.


\subsection{PASCAL VOC}
\label{sec:voc}

\begin{table}[t]
\small
\centering
\begin{tabular}{| c | c | c | c |}
\hline
PCK[$\alpha=0.1$] & Long\cite{long2014convnets} & VKps\cite{tulsiani2015viewpoints} & \  \ DISCO  \ \ \\
\hline
Full    			& 55.7 & 81.3 & \textbf{81.8} \\
Full[$\alpha=0.2$]  & NA & 88.3 & \textbf{93.4} \\
Occluded            & NA & \textbf{62.8} & 59.0 \\
Big Image     		& NA & \textbf{90.0} & 87.7 \\
Small Image    		& NA & 67.4 & \textbf{74.3} \\
\hhline{|=|=|=|=|}
All [APK $\alpha=0.1$] & NA & 40.3 & \textbf{45.4} \\
\hline
\end{tabular}
\caption{\small PCK[$\alpha=0.1$] accuracies (\%) of different methods for 2D keypoint localization on the car category of PASCAL VOC. Bold numbers indicate the best results.}
\label{tab:voc}
\end{table}

We evaluate DISCO on the PASCAL VOC 2012 dataset for 2D keypoint localization~\cite{yang2011articulated}. Unlike KITTI-3D where car images are captured on real roads and mostly in low resolution, PASCAL VOC contains car images with larger appearance variations and heavy occlusions. 
In Table, \ref{tab:voc}, we compare our results with state-of-the-art \cite{tulsiani2015viewpoints,long2014convnets} on various sub-classes of the test set: fully visible cars (denoted as ``Full''), occluded cars, high-resolution (average size $420$x$240$) and low-resolution images (average size $55$x$30$). Please refer to \cite{tulsiani2015viewpoints} for details of the test setup.

We observe that DISCO outperforms \cite{tulsiani2015viewpoints} by $0.6\%$ and $5.1\%$ on PCK at $\alpha=0.1$ and $\alpha=0.2$, respectively. In addition, DISCO is robust to low-resolution images, improving $6.9\%$ accuracy on low-resolution set compared with \cite{tulsiani2015viewpoints}. 
However, DISCO is inferior on the occluded car class and high-resolution images, attributable to our use of small images ($64$x$64$) for training 
and the fact that our occlusion simulation cannot capture more complex occlusion in typical road scenes. 
Finally, we compute APK accuracy at $\alpha=0.1$ for DISCO on the same detection candidates used in \cite{tulsiani2015viewpoints} \footnote{We run the source code provided by \cite{tulsiani2015viewpoints} to obtain the same object candidates.}.
We can see that DISCO outperforms \cite{tulsiani2015viewpoints} by $5.1\%$ on the entire car dataset (Full+Occluded).
This suggests DISCO is more robust to the noisy detection results and more accurate on keypoint visibility inference than \cite{tulsiani2015viewpoints}.
We attribute this to global structure modeling of DISCO during training where the full set of 2D keypoints teaches the network to resolve the partial view ambiguity.

Note that some definitions of our car keypoints~\cite{zia2013detailed} are slightly different from \cite{yang2011articulated}.
For example, we annotate the bottom corners of the front windshield but \cite{yang2011articulated} label the side mirrors. 
In our experiments, we ignore this annotation inconsistency and directly apply the prediction results.
Further, unlike \cite{long2014convnets,tulsiani2015viewpoints}, we do not use the PASCAL VOC train set, since our intent is to study the impact of deep supervision with shape concepts available through a rendering pipeline. Thus, even better performance is expected when real images with consistent labels are used for training.

\subsection{PASCAL3D+}
\label{sec:pascal3d}

\begin{table}[t]
\small
\centering
\begin{tabular}{| c | c | c |}
\hline
Method 									& CAD alignment GT 	& Manual GT \\
\hline
VDPM-16 \cite{xiang2014beyond}    		& NA   				& 51.9  \\
Xiang et al. \cite{mottaghi2015coarse} 	& 64.4 				& 64.3  \\
Random CAD \cite{xiang2014beyond}       & NA   				& 61.8  \\
GT CAD \cite{xiang2014beyond}   	 	& NA   				& 67.3  \\
DISCO    								& \textbf{71.2} & \textbf{67.6}  \\
\hline
\end{tabular}
\caption{\small Object segmentation accuracies (\%) of different methods on PASCAL3D+. Best results are shown in bold.}
\label{tab:pascal3d}
\end{table}

PASCAL3D+ \cite{xiang2014beyond} provides object viewpoint annotations for PASCAL VOC objects by aligning manually chosen 3D object CAD models onto the visible 2D keypoints. Because only a few CAD models are used for each category, 3D keypoint locations are not accurate. Thus, we use the evaluation metric proposed by \cite{xiang2014beyond} which measures the 2D segmentation accuracy\footnote{The standard IoU segmentation metric on PASCAL VOC benchmark.} of its projected model mask. With a 3D skeleton of an object, we are able to create a coarse object mesh based on the geometry and compute segmentation masks by projecting coarse mesh surfaces onto 2D image based on the estimated 2D keypoint locations. Please refer to the supplementary document for more details.

Table \ref{tab:pascal3d} reports the object segmentation accuracies on two types of ground truths. The column ``Manual GT'', is the manual pixel-level annotation provided by PASCAL VOC 2012, whereas ``CAD alignment GT'' uses the 2D projections of aligned CAD models as ground truth. Note that ``CAD alignment GT'' covers the entire object extent in the image including regions occluded by other objects. DISCO significantly outperforms the state-of-the-art method \cite{xiang2015data} by $4.6\%$ and $6.6\%$ using only synthetic data for training. Moreover, on ``Manual GT'' benchmark, we compare DISCO with ``Random CAD'' and ``GT CAD'' which stand for the projected segmentation of randomly selected and ground truth CAD models respectively, given the ground truth object pose. 
We find that DISCO yields even superior performance to ``GT CAD''.
This provides evidence that joint modeling of 3D geometry manifold and viewpoint is better than the pipeline of object retrieval plus alignment.
Further, we emphasize \emph{at least two orders of magnitude faster inference} of a forward pass of DISCO during testing compared with other sophisticated CAD alignment approaches.

\subsection{IKEA Dataset}
\label{sec:ikea}

\begin{table}[t]
\small
\centering
\begin{tabular}{| c | c | c || c | c |}
\hline
\multirow{2}{*}{Method}	&	\multicolumn{2}{c||}{Sofa} & \multicolumn{2}{c|}{Chair}\\
\hhline{|~|-|-||-|-|} & Avg. Recall & PCK & Avg. Recall & PCK  \\
\hline
3D-INN  & \textbf{88.0} & 31.0 & 87.8 & 41.4\\
DISCO  & 83.4 & \textbf{38.5} & \textbf{89.9} & \textbf{63.9}\\
\hline
\end{tabular}
\caption{\small Average recall and PCK[$\alpha=0.1$] accuracy(\%) for 3D structure prediction on the sofa and chair classes in IKEA dataset. }
\label{tab:ikea}
\end{table}

In this section, we evaluate DISCO on IKEA dataset \cite{lim2013parsing} with 3D keypoint annotations provided by \cite{wu2016single}.
We train a single DISCO network from scratch using 200K synthetic images of both chair and sofa instances, in order to evaluate whether DISCO is capable of learning multiple 3D object geometries simultaneously. 
At test time, we compare DISCO with the state-of-the-art 3D-INN\cite{wu2016single} on IKEA.
In order to remove the error on the viewpoint estimation for 3D structure evaluation as 3D-INN does, we compute the PCA bases of both the estimated 3D keypoints and their groundtruth. 
Next, we align two PCA bases and rotate the predicted 3D structure back to the canonical frame of the groundtruth.  
Table \ref{tab:ikea} reports the PCK[$\alpha=0.1$] and average recall\cite{wu2016single} (mean PCK over densely sampled $\alpha$ within $[0,1]$) of 3D-INN and DISCO on both sofa and chair classes.
We retrieve the PCK accuracy for 3D-INN from its publicly released results on IKEA dataset.
DISCO significantly outperforms 3D-INN on PCK, which means that DISCO obtains more correct predictions than 3D-INN.
This substantiates that direct exploitation of rich visual details from images adopted by DISCO is critical to infer more accurate and fine-grained 3D structure than lifting sparse 2D keypoints to 3D shapes like 3D-INN. 
However, DISCO is inferior to 3D-INN in terms of average recall on the sofa class.
This indicates that the wrong predictions by DISCO deviate more from the groundtruth than 3D-INN.
This is mainly because 3D predicted shapes from 3D-INN are constrained by shape bases so even wrong estimates have realistic object shapes when recognition fails. 
We conclude that DISCO is able to learn 3D patterns of object classes besides the car category and shows potential as a general-purpose approach to jointly model 3D geometric structures of multiple objects.
 
\subsection{Qualitative Results}
\label{sec:quality}

\begin{figure*}[t]
  \centering
    \includegraphics[width=1.0\linewidth]{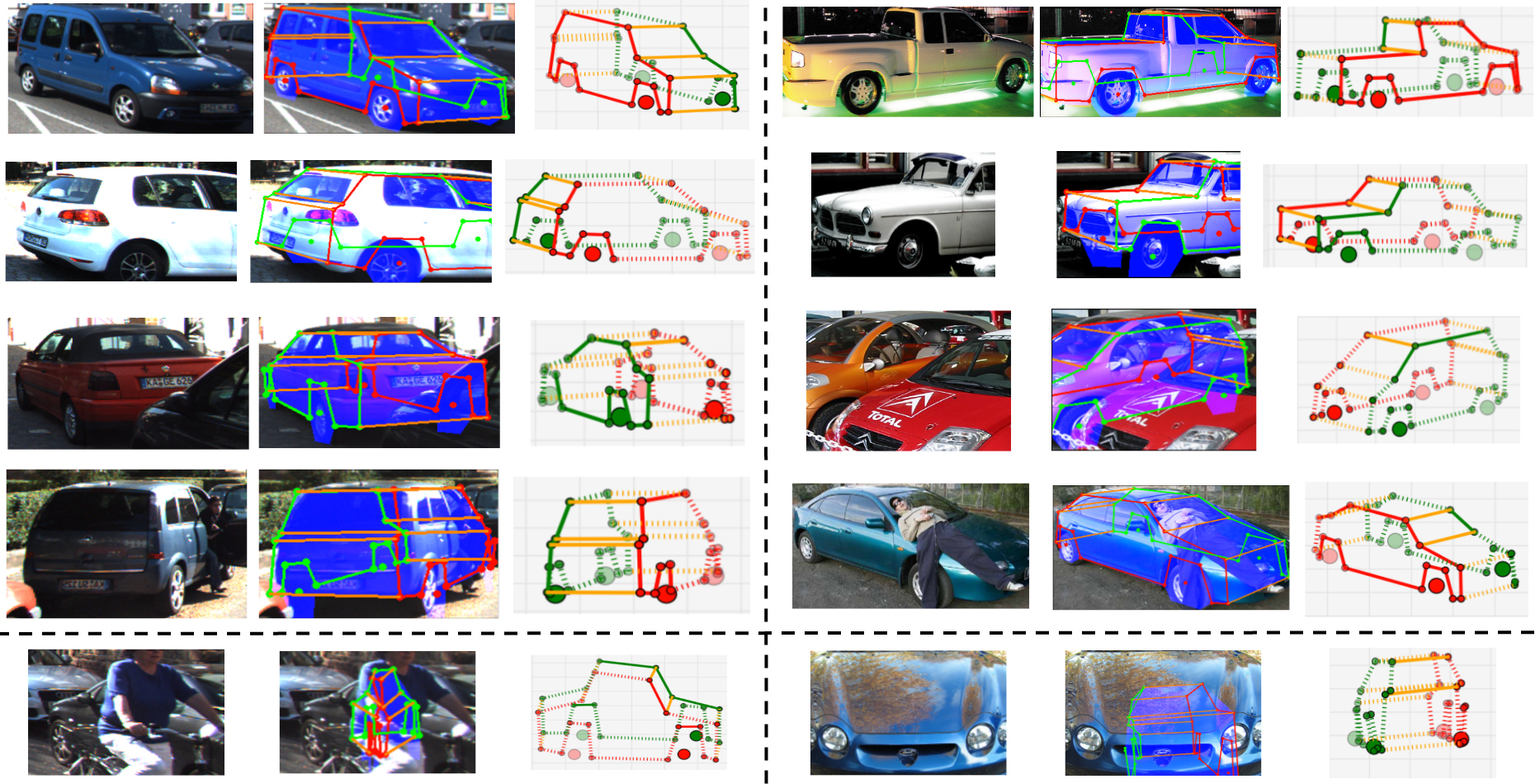}
  \caption{Visualization of 2D/3D prediction, visibility inference and instance segmentation on KITTI-3D (left column) and PASCAL VOC (right column). Last row shows failure cases. Circles and lines represent keypoints and their connections. Red and green indicate the left and right sides of a car, orange lines connect two sides. Dashed lines connect keypoints if one of them is inferred to be occluded. Light blue masks present segmentation results.
}
 \label{fig:demo}
\end{figure*}

\begin{figure*}[t]
  \centering
    \includegraphics[width=0.8\linewidth]{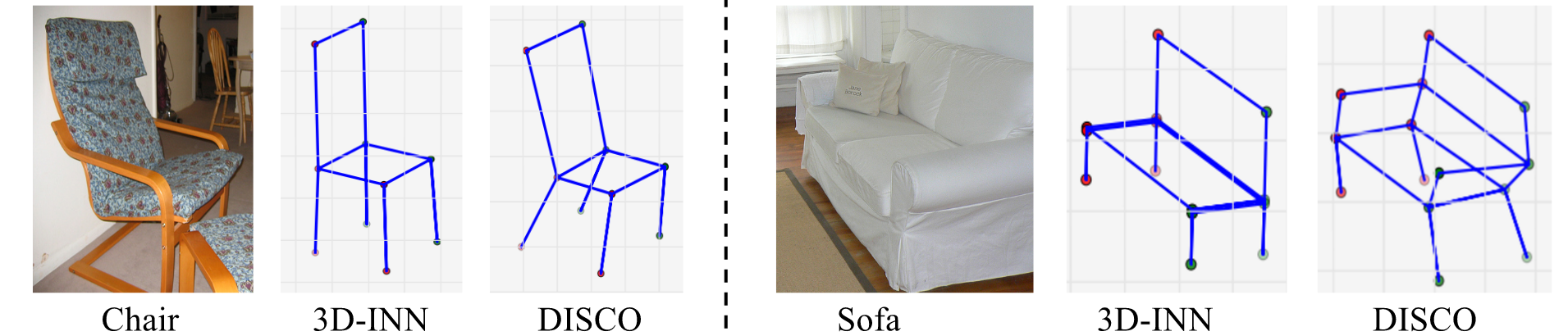}
  \caption{Qualitative comparison between 3D-INN and DISCO for 3D stricture prediction on IKEA dataset.}
 \label{fig:ikea_demo}
\end{figure*}

In Figure \ref{fig:demo}, we demonstrate example predictions from DISCO on KITTI-3D and PASCAL VOC. From left to right, each row shows the original object image, the predicted 2D object skeleton as well as instance segmentation and 3D object skeleton with visibility. 
We visualize example results under no occlusion (rows 1), truncation (row 2), multi-car occlusion (row 3) and other occluders (row 4). 
We can see that DISCO can localize 2D and 3D keypoints on real images with complex occlusion scenarios and diverse car models such as sedan, SUV and pickup. 
Moreover, visibility inference by DISCO is mostly correct. 
These capabilities highlight the potential of DISCO as a building block for holistic scene understanding in cluttered scenes. 
The last row shows two failure cases where the left car is mostly occluded by another object and the right one is severely truncated and distorted in projection. 
We may improve the performance of DISCO on these challenging cases by training DISCO on both synthetic data simulated with more complex occlusions~\cite{richter2016playing} and real data with 2D and 3D annotations.

Finally, we qualitatively compare 3D-INN and DISCO on two examples visualized in Fig. \ref{fig:ikea_demo}.
In the chair example, 3D-INN fails to delineate the inclined seatback.
For the sofa, DISCO captures the sofa armrest whereas 3D-INN merges armrests to the seating area.
We attribute this relative success of DISCO to the direct mapping from image to 3D structure, as opposed to lifting 2D keypoint predictions to 3D.

\section{Conclusion}
\label{sec:con}

We present a framework that deeply supervises a CNN architecture to incrementally develop 2D/3D shape understanding using a series of intermediate shape concepts. 
A 3D CAD model rendering pipeline generates numerous synthetic training images with supervisory signals for the deep supervision. 
The fundamental relationship of the shape concepts to 3D reconstruction is supported by our network generalizing well to real images at test time, despite our synthetic renderings not being photorealistic. 
Experiments demonstrate that our network outperforms current state-of-the-art methods on 2D and 3D landmark prediction on public datasets, even with occlusion and truncation. 
Further, we present preliminary results on jointly learning 3D geometry of multiple object classes within a single CNN.
Our future work will extend this direction by learning representations for diverse object classes. 
The present method is unable to model highly deformable objects due to lack of CAD training data, and topologically inconsistent object categories such as buildings.
These are also avenues for future work. 
More interestingly, our deep supervision can potentially be applied to tasks with abundant intermediate concepts such as scene physics inference.


\section*{Acknowledgments}
This work was part of C. Li's intern project at NEC Labs America, in Cupertino. We also acknowledge the support by NSF under Grant No. NRI-1227277.

{\small
\bibliographystyle{ieee}
\bibliography{ref}

\begin{thebibliography}{10}\itemsep=-1pt

\bibitem{aubry2014seeing}
M.~Aubry, D.~Maturana, A.~Efros, B.~Russell, and J.~Sivic.
\newblock {Seeing 3D chairs: Exemplar part-based 2D-3D alignment using a large
  dataset of CAD models}.
\newblock In {\em CVPR}, 2014.

\bibitem{bansal2016marr}
A.~Bansal, B.~Russell, and A.~Gupta.
\newblock {Marr revisited: 2D-3D Alignment via Surface Normal Prediction}.
\newblock In {\em CVPR}, 2016.

\bibitem{chang2015shapenet}
A.~X. Chang, T.~Funkhouser, L.~Guibas, P.~Hanrahan, et~al.
\newblock Shapenet: An information-rich 3d model repository.
\newblock {\em arXiv:1512.03012}, 2015.

\bibitem{choy20163d}
C.~B. Choy, D.~Xu, J.~Gwak, K.~Chen, and S.~Savarese.
\newblock {3D-R2N2: A Unified Approach for Single and Multi-view 3D Object
  Reconstruction}.
\newblock In {\em ECCV}, 2016.

\bibitem{dosovitskiy2016learning}
A.~Dosovitskiy, J.~Springenberg, and T.~Brox.
\newblock {Learning to Generate Chairs with Convolutional Neural Networks}.
\newblock In {\em CVPR}, 2015.

\bibitem{geiger2012we}
A.~Geiger, P.~Lenz, and R.~Urtasun.
\newblock {Are we ready for Autonomous Driving? The KITTI Vision Benchmark
  Suite}.
\newblock In {\em CVPR}, 2012.

\bibitem{girshick2014rich}
R.~Girshick, J.~Donahue, T.~Darrell, and J.~Malik.
\newblock Rich feature hierarchies for accurate object detection and semantic
  segmentation.
\newblock In {\em CVPR}, 2014.

\bibitem{glorot2010understanding}
X.~Glorot and Y.~Bengio.
\newblock {Understanding the difficulty of training deep feedforward neural
  networks}.
\newblock {\em {AISTATS}}, 2010.

\bibitem{gupta2015inferring}
S.~Gupta, P.~Arbel{\'a}ez, R.~Girshick, and J.~Malik.
\newblock {Inferring 3d object pose in RGB-D images}.
\newblock {\em arXiv:1502.04652}, 2015.

\bibitem{ioffe2015batch}
S.~Ioffe and C.~Szegedy.
\newblock {Batch Normalization: Accelerating Deep Network Training by Reducing
  Internal Covariate Shift}.
\newblock {\em JMLR}, 2015.

\bibitem{jia2014caffe}
Y.~Jia, E.~Shelhamer, J.~Donahue, S.~Karayev, J.~Long, R.~Girshick,
  S.~Guadarrama, and T.~Darrell.
\newblock Caffe: Convolutional architecture for fast feature embedding.
\newblock {\em arXiv:1408.5093}, 2014.

\bibitem{kanazawa2016warpnet}
A.~Kanazawa, D.~W. Jacobs, and M.~Chandraker.
\newblock {WarpNet: Weakly Supervised Matching for Single-view Reconstruction}.
\newblock In {\em CVPR}, 2016.

\bibitem{kar2015category}
A.~Kar, S.~Tulsiani, J.~Carreira, and J.~Malik.
\newblock Category-specific object reconstruction from a single image.
\newblock In {\em CVPR}, 2015.

\bibitem{krizhevsky2012imagenet}
A.~Krizhevsky, I.~Sutskever, and G.~E. Hinton.
\newblock {ImageNet Classification with Deep Convolutional Neural Networks}.
\newblock In {\em NIPS}, 2012.

\bibitem{kulkarni2014inverse}
T.~D. Kulkarni, W.~F. Whitney, P.~Kohli, and J.~B. Tenenbaum.
\newblock Deep convolutional inverse graphics network.
\newblock In {\em NIPS}, 2015.

\bibitem{lee2015deeply}
C.-Y. Lee, S.~Xie, P.~Gallagher, Z.~Zhang, and Z.~Tu.
\newblock {Deeply-Supervised Nets}.
\newblock {\em AISTATS}, 2015.

\bibitem{lee1985determination}
H.-J. Lee and Z.~Chen.
\newblock {Determination of 3D human body postures from a single view}.
\newblock {\em CVGIP}, 1985.

\bibitem{lim2014fpm}
J.~J. Lim, A.~Khosla, and A.~Torralba.
\newblock {FPM: Fine pose Parts-based Model with 3D CAD models}.
\newblock In {\em ECCV}, 2014.

\bibitem{lim2013parsing}
J.~J. Lim, H.~Pirsiavash, and A.~Torralba.
\newblock {Parsing IKEA Objects: Fine Pose Estimation}.
\newblock In {\em ICCV}, 2013.

\bibitem{long2014convnets}
J.~L. Long, N.~Zhang, and T.~Darrell.
\newblock Do convnets learn correspondence?
\newblock In {\em NIPS}, 2014.

\bibitem{Lowe_1985}
D.~G. Lowe.
\newblock {\em Perceptual Organization and Visual Recognition}.
\newblock Kluwer Academic Publishers, Norwell, MA, USA, 1985.

\bibitem{Marr_1982}
D.~Marr.
\newblock {\em Vision}.
\newblock Henry Holt and Co., Inc., 1982.

\bibitem{massa2015deep}
F.~Massa, B.~Russell, and M.~Aubry.
\newblock {Deep Exemplar 2D-3D Detection by Adapting from Real to Rendered
  Views}.
\newblock In {\em CVPR}, 2015.

\bibitem{Mohan_Nevatia_1989}
R.~Mohan and R.~Nevatia.
\newblock Using perceptual organization to extract 3{D} structures.
\newblock {\em PAMI}, 1989.

\bibitem{pol2016overcoming}
P.~Moreno, C.~K. Williams, C.~Nash, and P.~Kohli.
\newblock Overcoming occlusion with inverse graphics.
\newblock In {\em ECCV}, 2016.

\bibitem{mottaghi2015coarse}
R.~Mottaghi, Y.~Xiang, and S.~Savarese.
\newblock A coarse-to-fine model for 3d pose estimation and sub-category
  recognition.
\newblock In {\em CVPR}, 2015.

\bibitem{pepik13cvpr}
B.~Pepik, M.~Stark, P.~Gehler, and B.~Schiele.
\newblock Occlusion patterns for object class detection.
\newblock In {\em CVPR}, 2013.

\bibitem{rezende2016unsupervised}
D.~J. Rezende, S.~Eslami, S.~Mohamed, P.~Battaglia, M.~Jaderberg, and N.~Heess.
\newblock Unsupervised learning of 3d structure from images.
\newblock In {\em NIPS}, 2016.

\bibitem{richter2016playing}
S.~R. Richter, V.~Vineet, S.~Roth, and V.~Koltun.
\newblock Playing for data: Ground truth from computer games.
\newblock In {\em ECCV}, 2016.

\bibitem{Sarkar_Soundararajan_2000}
S.~Sarkar and P.~Soundararajan.
\newblock Supervised learning of large perceptual organization: {G}raph
  spectral partitioning and learning automata.
\newblock {\em PAMI}, 2000.

\bibitem{simonyan2014very}
K.~Simonyan and A.~Zisserman.
\newblock Very deep convolutional networks for large-scale image recognition.
\newblock {\em arXiv:1409.1556}, 2014.

\bibitem{Smith_1986}
B.~J. Smith.
\newblock {\em {Perception of Organization in a Random Stimulus}}.
\newblock 1986.

\bibitem{su2015render}
H.~Su, C.~R. Qi, Y.~Li, and L.~J. Guibas.
\newblock {Render for CNN: Viewpoint estimation in images using CNNs trained
  with Rendered 3D model views}.
\newblock In {\em ICCV}, 2015.

\bibitem{tatarchenko2016multi}
M.~Tatarchenko, A.~Dosovitskiy, and T.~Brox.
\newblock {Multi-view 3D Models from Single Images with a Convolutional
  Network}.
\newblock In {\em ECCV}, 2016.

\bibitem{torresani2003learning}
L.~Torresani, A.~Hertzmann, and C.~Bregler.
\newblock Learning non-rigid 3d shape from 2d motion.
\newblock In {\em Advances in Neural Information Processing Systems}, page
  None, 2003.

\bibitem{tulsiani2015viewpoints}
S.~Tulsiani and J.~Malik.
\newblock {Viewpoints and Keypoints}.
\newblock In {\em CVPR}, 2015.

\bibitem{wu2016single}
J.~Wu, T.~Xue, J.~J. Lim, Y.~Tian, J.~B. Tenenbaum, A.~Torralba, and W.~T.
  Freeman.
\newblock {Single Image 3D Interpreter Network}.
\newblock In {\em ECCV}, 2016.

\bibitem{wu2015learning}
T.~Wu, B.~Li, and S.-C. Zhu.
\newblock {Learning And-Or Model to Represent Context and Occlusion for Car
  Detection and Viewpoint Estimation}.
\newblock {\em PAMI}, 2016.

\bibitem{xiang2015data}
Y.~Xiang, W.~Choi, Y.~Lin, and S.~Savarese.
\newblock {Data-driven 3D voxel patterns for object category recognition}.
\newblock In {\em CVPR}, 2015.

\bibitem{xiang2014beyond}
Y.~Xiang, R.~Mottaghi, and S.~Savarese.
\newblock {Beyond PASCAL: A Benchmark for 3D Object Detection in the Wild}.
\newblock In {\em WACV}, 2014.

\bibitem{yang2011articulated}
Y.~Yang and D.~Ramanan.
\newblock Articulated pose estimation with flexible mixtures-of-parts.
\newblock In {\em CVPR}, 2011.

\bibitem{yu2016deep}
X.~Yu, F.~Zhou, and M.~Chandraker.
\newblock {Deep Deformation Network for Object Landmark Localization}.
\newblock {\em ECCV}, 2016.

\bibitem{zhou2016learning}
T.~Zhou, P.~Kr{\"a}henb{\"u}hl, M.~Aubry, Q.~Huang, and A.~A. Efros.
\newblock {Learning Dense Correspondence via 3D-guided Cycle Consistency}.
\newblock In {\em CVPR}, 2016.

\bibitem{zia09icar}
M.~Z. Zia, U.~Klank, and M.~Beetz.
\newblock {Acquisition of a Dense 3D Model Database for Robotic Vision}.
\newblock In {\em ICAR}, 2009.

\bibitem{zia2013detailed}
M.~Z. Zia, M.~Stark, and K.~Schindler.
\newblock {Towards Scene Understanding with Detailed 3D Object
  Representations}.
\newblock {\em IJCV}, 2015.

\end{thebibliography}
}


\end{document}